\documentclass[10pt,twocolumn,letterpaper]{article}

\usepackage[pagenumbers]{cvpr}

\usepackage{graphicx}
\usepackage{amsmath}
\usepackage{amssymb}
\usepackage{booktabs}
\usepackage[misc]{ifsym}
\usepackage{times}
\usepackage{epsfig}
\usepackage{multirow}
\usepackage{soul}
\usepackage[accsupp]{axessibility}  
\usepackage{xspace}
\usepackage{mathrsfs}
\usepackage{soul}
\usepackage{array}
\usepackage{tabu}
\usepackage{color,xcolor}
\usepackage{colortbl}
\usepackage{threeparttable}
\definecolor{mygray}{gray}{.85}
\definecolor{mygray1}{gray}{.7}
\definecolor{mygray2}{gray}{.93}
\newcommand{\myparagraph}[1]{{\vspace{.5em} \noindent \bf #1}}
\usepackage{makecell}
\usepackage{flushend}

\newcommand{\numvideo}{2,006\xspace}
\newcommand{\numobject}{8,171\xspace}

\newcommand{\numsentence}{28,570\xspace}
\newcommand{\ourdataset}{MeViS\xspace}

\newcommand{\fullname}{\textbf{M}otion \textbf{e}xpressions \textbf{Vi}deo \textbf{S}egmentation\xspace}

\newcommand{\ourmodel}{LMPM\xspace}

\usepackage{pifont}
\newcommand{\cmark}{\ding{51}\xspace}%
\newcommand{\xmarkg}{\textcolor{lightgray}{\ding{55}}\xspace}%

\makeatletter
\newcommand{\thickhline}{%
    \noalign {\ifnum 0=`}\fi \hrule height 1pt
    \futurelet \reserved@a \@xhline
}
\makeatother

\usepackage[pagebackref,breaklinks,colorlinks]{hyperref}

\usepackage[capitalize]{cleveref}
\crefname{section}{Sec.}{Secs.}
\Crefname{section}{Section}{Sections}
\Crefname{table}{Table}{Tables}
\crefname{table}{Tab.}{Tabs.}

\begin{document}

\title{MeViS: A Large-scale Benchmark for Video Segmentation \\with Motion Expressions}

\author{
Henghui Ding
\quad
Chang Liu
\quad
Shuting He
\quad
Xudong Jiang
\quad
Chen Change Loy\\
Nanyang Technological University
\\
\href{https://henghuiding.github.io/MeViS}{https://henghuiding.github.io/MeViS}\\
}

\twocolumn[{%
\renewcommand\twocolumn[1][]{#1}%
\maketitle
\begin{center} 
\centering 
\vspace{-1.6mm}
\captionsetup{type=figure}
\includegraphics[width=0.999\textwidth]{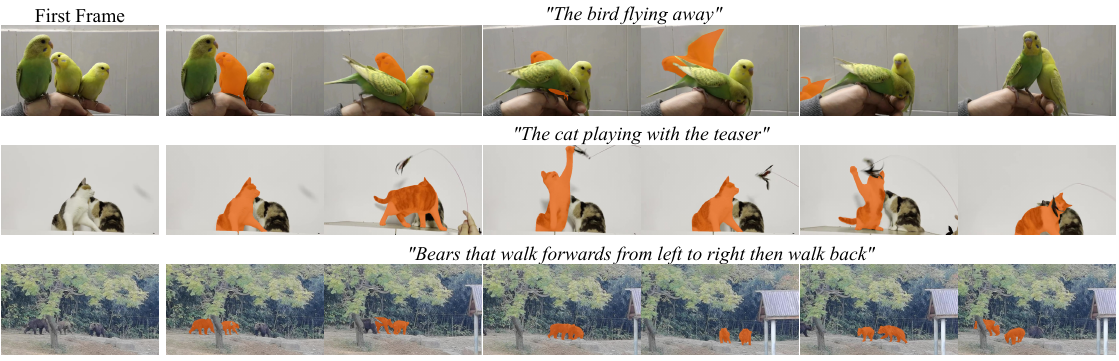}
\vspace{-6.6mm}
\captionof{figure}{\small Examples of video clips from \fullname (\textbf{\ourdataset}) are provided to illustrate the dataset's nature and complexity. The expressions in \ourdataset primarily focus on motion attributes and the referred target object cannot be identified by examining a single frame solely.  For instance, the first example features three parrots with similar appearances, and the target object is identified as \textit{``The bird flying away''}. This object can only be recognized by capturing its motion throughout the video.
}
\vspace{1.6mm}
\label{Fig:teaser}
\end{center}
}]

\renewcommand{\thefootnote}{\fnsymbol{footnote}}
\footnotetext[0]{${\textrm{\Letter}}$ henghui.ding@gmail.com, ccloy@ntu.edu.sg}

\begin{abstract}
\vspace{-1.6mm}
   This paper strives for motion expressions guided video segmentation, which focuses on segmenting objects in video content based on a sentence describing the motion of the objects. Existing referring video object datasets typically focus on salient objects and use language expressions that contain excessive static attributes that could potentially enable the target object to be identified in a single frame. These datasets downplay the importance of motion in video content for language-guided video object segmentation. To investigate the feasibility of using motion expressions to ground and segment objects in videos, we propose a large-scale dataset called \ourdataset, which contains numerous motion expressions to indicate target objects in complex environments. We benchmarked 5 existing referring video object segmentation (RVOS) methods and conducted a comprehensive comparison on the \ourdataset dataset. The results show that current RVOS methods cannot effectively address motion expression-guided video segmentation. We further analyze the challenges and propose a baseline approach for the proposed \ourdataset dataset. The goal of our benchmark is to provide a platform that enables the development of effective language-guided video segmentation algorithms that leverage motion expressions as a primary cue for object segmentation in complex video scenes. The proposed MeViS dataset has been released at \href{https://henghuiding.github.io/MeViS}{https://henghuiding.github.io/MeViS}.
\end{abstract}

\section{Introduction}
\label{sec:intro}
Language-guided video segmentation is an emerging field that involves segmenting and tracking target objects using natural language expressions. This field has traditionally been a sub-branch of semi-supervised video object segmentation, where referring expressions are used to describe the target object. Existing referring video object datasets, such as~\cite{gavrilyuk2018actor,khoreva2018video,seo2020urvos}, commonly feature videos with isolated and salient objects that have obvious static features. The corresponding expressions often contain static attributes such as object color, which can be observed in a single frame. As a result, motion properties of videos are often given less emphasis, and referring image segmentation methods can be used for referring video segmentation, achieving good results \cite{khoreva2018video,vltpami,bellver2020refvos,liu2021cmpc}.

In this paper, we wish to highlight the significance of temporal motion properties of videos and explore the potential of using motion expressions to segment objects in videos. To this end, we propose a new large-scale dataset called \textbf{M}otion \textbf{e}xpressions \textbf{Vi}deo \textbf{S}egmentation (\textbf{\ourdataset}) to aid our investigation. The \ourdataset dataset comprises \numvideo videos with a total of \numobject objects, and \numsentence motion expressions are provided to refer to these objects.

We take several steps to ensure that the \ourdataset dataset places emphasis on the temporal motions of videos. First, we carefully select video content that contains multiple objects that coexist with motion and exclude videos with isolated objects that can be easily described by static attributes. Second, we prioritize language expressions that do not contain static clues, such as category names or object colors, in cases where target objects can be unambiguously described by motion words alone. This is distinct from previous datasets, such as~\cite{gavrilyuk2018actor,khoreva2018video,seo2020urvos}, which include obvious static clues in their expressions. Additionally, \ourdataset differentiates itself from referring image segmentation datasets, such as~\cite{yu2016modeling,mao2016generation,wu2020phrasecut,kazemzadeh-etal-2014-referitgame}, which do not account for the temporal properties of video content. Moreover, unlike existing referring video object segmentation datasets that focus on single-target expressions, where one expression refers to only one target object, \ourdataset expands this task to include multi-object expressions that refer to multiple target objects. This feature enables expressions to refer to an unlimited number of target objects, making the proposed \ourdataset more challenging and reflective of real-world scenarios.

The proposed \ourdataset dataset poses notable challenges in capturing and understanding motions in both video and language. The language expressions may describe motion that spans a random number of frames, requiring the capture of fleeting movements and long-term actions that occur throughout the entire video. This poses significant challenges for both understanding motion in the video content and in the accompanying language expressions. Capturing fleeting movements requires attention on each individual frame, while understanding long and complex movements that span across many frames demands temporal context across the entire video. With the proposed dataset, we benchmark 5 existing referring video object segmentation (RVOS) methods~\cite{vltpami,wu2022referformer,MTTR,Ding_2022_CVPR,seo2020urvos} and conduct a comprehensive comparison. The experimental results demonstrate that \ourdataset presents more challenges than existing datasets, and current RVOS methods are unable to effectively address motion expression-guided video segmentation.

In addition to proposing the \ourdataset dataset, we present a baseline approach, named Language-guided Motion Perception and Matching (\ourmodel), to address the challenges posed by the dataset. Our approach generates language-conditional queries to detect potential target objects in the video and represents them using object embeddings, which are more robust and computationally efficient than object feature maps~\cite{VITA}. We then perform Motion Perception on the object embeddings to capture the temporal context and obtain a global view of the video, enabling the model to understand both fleeting and long-term motions. Next, we use a Transformer decoder to decode language-related information from the motion-aggregated object embeddings and predict object trajectories. 
Finally, we perform similarity matching between the language features and the predicted object trajectories to identify the target object(s).

Our contributions provide a foundation for developing more advanced language-guided video segmentation algorithms that leverage motion expressions as a primary cue for object segmentation and identification in complex video scenes. In particular, we propose a new language-guided video segmentation dataset, \textbf{\ourdataset}, and conduct comprehensive evaluations of state-of-the-art referring video object segmentation methods on the \ourdataset dataset, providing a reference for future works. We also develop a simple baseline approach, \ourmodel, which points to potential solutions to some of the challenges and future research directions.

\section{Related Work}

\noindent\textbf{Referring Image Segmentation.}
Referring image segmentation~\cite{ding2021vision,vltpami,GRES,ding2020phraseclick,M3Att}, also known as referring expression segmentation, involves grounding the target object in images based on natural language expressions that describe its properties and generating a corresponding segmentation mask. This task requires both language and image understanding and is one of the most fundamental yet challenging tasks in computer vision. Referring image segmentation was first introduced by Hu~\etal~\cite{hu2016segmentation} in 2016 and has received consideratble attention since then. In the pre-Transformer era, mainstream methods typically employed Fully Convolutional Networks (FCN)~\cite{long2015fully,ding2018context,BFP} and Recurrent Neural Networks (RNN) to extract image features and language features, respectively, and then fused the multi-modal features via some specially designed modules~\cite{liu2017recurrent,li2018referring,feng2021encoder,margffoy2018dynamic}. For example, Liu \etal~\cite{liu2017recurrent} introduced a Recurrent Multimodal Interaction (RMI) module to recurrently fuse the feature of each word into the image features. Li \etal~\cite{li2018referring} proposed a Recurrent Refinement Network (RRN) that progressively refines the segmentation mask based on pyramid features in FCN.

In addition to one-stage methods that fuse multi-modal features and conduct segmentation, some methods decouple referring image segmentation into instance segmentation and language-object matching~\cite{yu2018mattnet, ISFP, jing2021locate}. For instance, Yu \etal use the off-the-shelf instance segmentation model Mask R-CNN~\cite{he2017mask} to detect all instances first and then select the one that best matches the language as output. A holistic understanding of language and vision information is crucial for referring image segmentation, and many works have explored this direction~\cite{hui2020linguistic,yang2021bottom,ye2019cross}. For instance, Ye \etal introduce a Cross-Modal Self-Attention (CMSA) model~\cite{ye2019cross} to select the most meaningful words in the expression and pixels in the image to achieve better contextual understanding. Recently, the success of Transformer~\cite{vaswani2017attention} in vision tasks has inspired many studies in referring image segmentation. Ding \etal~\cite{vltpami,ding2021vision} first introduced Transformer into referring segmentation and proposed a Vision-Language Transformer (VLT). Following Ding \etal~\cite{vltpami,ding2021vision}, more Transformer-based methods have been proposed~\cite{yang2021lavt, wang2022cris, kim2022restr}. For example, Wang \etal~\cite{wang2022cris} employ the Vision-Language Decoder to deal with visual and text tokens extracted by CLIP~\cite{radford2021learning}. Yang \etal~\cite{yang2021lavt} focus on multi-modal feature fusion and propose a Language-Aware Vision Transformer (LAVT).

\noindent\textbf{Referring Video Segmentation.} Referring video object segmentation is an emerging area~\cite{wang2020context,ningpolar,wang2019asymmetric,mcintosh2020visual,liu2021cmpc,hui2021collaborative,Wu_2022_CVPR,zhao2022modeling,sun2022starting,chen2022multi,yang2022tubedetr,tang2021human} that aims to segment the target object indicated by a given expression across the entire video clip. It was first introduced in 2018 by A2D~\cite{gavrilyuk2018actor} and DAVIS${17}$-RVOS~\cite{khoreva2018video}, where A2D~\cite{gavrilyuk2018actor} seeks to segment actors according to descriptions of their actions in video content, and DAVIS${17}$-RVOS~\cite{khoreva2018video} replaces masks with language as the reference for the target object in video object segmentation. Later, Seo \etal~\cite{seo2020urvos} built the Refer-YouTube-VOS based on the YouTube-VOS-2019 dataset~\cite{xu2018youtube}. These datasets typically provide an expression for a single object, and the expression usually describes the static attributes of the target object, such as its color and shape.

Existing methods typically treat referring video segmentation as a form of semi-supervised video object segmentation~\cite{davis2017} by replacing mask reference with language reference. For instance, Khoreva \etal~\cite{khoreva2018video} employ the referring image segmentation method MAttNet~\cite{yu2018mattnet} to achieve frame-level segmentation and then perform post-processing for temporal consistency. URVOS~\cite{seo2020urvos} employs cross-modal attention to perform per-frame segmentation and propagate the mask across clips with a memory attention module. RefVOS~\cite{bellver2020refvos} independently segments each frame based on the fused features of language and image/frame, without utilizing temporal information. Liang \etal~\cite{liang2021topdown} introduces a top-down approach that first detects all object tracklets and then selects the target object by matching between language and tracklet features. Most recently, ReferFormer~\cite{wu2022referformer} and MTTR~\cite{MTTR} employ Transformer~\cite{vaswani2017attention} to address referring video object segmentation.
\section{\ourdataset Dataset}\label{sec:MeViS_Dataset}
In this section, we introduce the newly built large-scale dataset \textbf{\ourdataset} by first presenting the video collection and annotation process in \Cref{sec:videoannotation} and then providing the dataset statistics and analysis in \Cref{sec:datasetStatistics}.

\begin{table*}
\centering
\small
\caption{Statistics of representative language-guided video segmentation datasets. The newly built \ourdataset has the largest number of objects and language expressions. More importantly, \ourdataset focuses on segmenting objects in the videos indicated by motion expressions. The \ourdataset enables the investigation of the feasibility of using motion expressions for object segmentation and grounding in videos.}
\vspace{-8pt}
\begin{threeparttable}
\setlength\tabcolsep{7.6pt}
\renewcommand\arraystretch{1.16}
\begin{tabular}{lccccccccc}
\hline
Dataset~~~~~~~~~&Year &Pub. &Video&Object &Expression & Mask & \makecell[c]{Object/\\Video} &\makecell[c]{Object/\\Experission}  &Target  \\
\hline
\hline
A2D Sentence~\cite{gavrilyuk2018actor} &2018 &CVPR &\makebox[5ex][r]{3,782}&\makebox[5ex][r]{4,825} &\makebox[6ex][r]{6,656} & 58k & 1.28 & 1 &Actor\\
J-HMDB Sentence~\cite{gavrilyuk2018actor}&2018 &CVPR  &\makebox[5ex][r]{928}&\makebox[5ex][r]{928} &\makebox[6ex][r]{928} &31.8k& 1 & 1 &Actor\\
DAVIS$_{16}$-RVOS~\cite{khoreva2018video} &2018 &ACCV  &\makebox[5ex][r]{50}& \makebox[5ex][r]{50}& \makebox[6ex][r]{100} &3.4k & 1 &n/a &Object\\
DAVIS$_{17}$-RVOS~\cite{khoreva2018video} &2018 &ACCV  &\makebox[5ex][r]{90}& \makebox[5ex][r]{205}& \makebox[6ex][r]{1,544} &13.5k & 2.27 & 1 &Object\\
Refer-Youtube-VOS~\cite{seo2020urvos} &2020 &ECCV  &\makebox[5ex][r]{\textbf{3,978}}& \makebox[5ex][r]{7,451}& \makebox[6ex][r]{15,009}& 131k & 1.86 & 1 &Object\\\hline
\textbf{\ourdataset} (ours) &2023 &ICCV & \numvideo&  \textbf{\numobject}&\textbf{\numsentence} & \textbf{443k} & \textbf{4.28} & \textbf{1.59} &Object(\textbf{s})\\
\hline
\end{tabular}
\end{threeparttable}
\label{table:dataset}
\vspace{-3.6mm}
\end{table*}
\subsection{Motion Expression Annotation}\label{sec:videoannotation}

\myparagraph{Video Collection.} We gather and choose videos from publicly available video segmentation datasets with high-quality mask annotations~\cite{OVIS,UVO,TAOVOS,MOSE}, and select the ones that meet our criteria for motion and object complexity. Our selection process involves the following rules:
\begin{itemize}
\setlength\itemsep{0em}
\vspace{-1mm}
\item[R1.] We only include videos that have multiple objects within the frame in \ourdataset; videos with only one or two salient objects are not considered. We specifically look for videos that depict many objects with similar appearances, such as the first example video in \Cref{Fig:teaser} which shows three yellow parrots.
\vspace{-1mm}

\item[R2.] We select videos that contain objects that demonstrate substantial motion and movement. Videos depicting objects that have little or no motion are excluded.
\vspace{-1mm}
\end{itemize}
\vspace{-1mm}

After reviewing over 4,000 potential candidates, we carefully selected the most appropriate and suitable videos that meet our rigorous standards for both visual and linguistic content. Ultimately, by prioritizing quality over quantity, we chose \numvideo videos to create a benchmark that is diverse and representative of a wide range of real-world video scenarios. The basic language annotation methodology and procedure for \ourdataset follow the ReferIt~\cite{kazemzadeh-etal-2014-referitgame}, which is an interactive game-like approach that involves two players taking turns to annotate and validate. The following section will introduce the process of language expression annotation and validation in more detail.

\myparagraph{Language Expression Annotation.} 
We developed a web-based annotation system for annotating language expressions. The system randomly selects a video from the \ourdataset dataset and displays all object masks of the selected video on the web system. The annotator needs to choose one or several objects from the video and write the corresponding referring expression according to the guidelines for annotating language expressions. To ensure that the language expressions in our dataset align with our focus on motion-based video segmentation, we established several guidelines for annotating the language expressions:
\begin{itemize}
\setlength\itemsep{0em}
\item[A1.] Target objects must exhibit significant motion. Objects that remain stationary or only demonstrate minimal motion should be disregarded.
\vspace{-1mm}
\item[A2.] If an object can be unambiguously described by its motion or action, static attributes such as color should not be included in the expression.
\vspace{-1mm}
\item[A3.] If multiple objects cannot be differentiated based solely on their motion or action, they can be described together if their motion or action can unambiguously identify them, such as ``\textit{The two lions fighting and running amidst a group of lions.}"
\vspace{-1mm}
\item[A4.] If it is not possible to differentiate single or multiple objects based solely on their motion or action, limited static attributes can be included in the expression.
\end{itemize}

\vspace{-3.6mm}
\myparagraph{Language Expression Validation.} 
Upon receiving annotated ``video-object-expression'' samples from the annotators, the validation process begins by displaying the video and expression and prompting the validator to select and submit the objects referred to in the expression. The validator must find the targets independently and submit their selection. The system then compares the targets chosen by the validator with the annotations submitted by the annotator. A sample is considered valid if the validator and annotator independently selected the same target object(s) using the same expression. If the targets selected by the validator do not match the annotation submitted by the annotator, the sample will be forwarded to another validator for a second opinion. If the second validator also fails to identify the correct targets, the sample will be considered invalid and excluded from the dataset. Validators have the authority to reject samples that are deemed inappropriate or fall short of quality standards. Moreover, we stress the importance of the following validation criterion:
\begin{itemize}
\vspace{-1.6mm}
\item[V1.] The corresponding sentence will be removed from the dataset when the target object described by a sentence can be identified through a single frame without the need for motion information.
\vspace{-1.6mm}
\end{itemize}
By establishing these validation criteria, we aim to ensure that the language sentences in our dataset accurately express motion and are of high quality, while also increasing the level of difficulty in the language-guided video segmentation task, thereby enabling a more robust evaluation of the performance of different models and methods.

\subsection{Dataset Analysis and Statistics}\label{sec:datasetStatistics}

In~\Cref{table:dataset}, we present a statistical analysis of the newly proposed \textbf{\ourdataset} dataset, using 5 previous referring video object segmentation datasets as references, including A2D Sentence~\cite{gavrilyuk2018actor}, J-HMDB Sentence~\cite{gavrilyuk2018actor}, DAVIS${16}$-RVOS~\cite{khoreva2018video}, DAVIS${17}$-RVOS~\cite{khoreva2018video}, and Refer-Youtube-VOS~\cite{seo2020urvos}. As shown in \Cref{table:dataset}, \ourdataset contains \numvideo videos and \numobject objects. Compared to Refer-Youtube-VOS~\cite{seo2020urvos}, which is based on the existing VOS dataset~\cite{xu2018youtube}, \ourdataset has more objects (\numobject \vs~7,451), more expressions (\numsentence \vs~15,009), and more annotation masks (443k \vs~131k). In the following, we discuss how the proposed dataset \ourdataset intentionally increases the complexities of language-guided video segmentation by considering the challenges of both linguistic and visual modalities.

\begin{figure}
    \centering
    \hspace{-5mm}
    \includegraphics[width=0.25\textwidth]{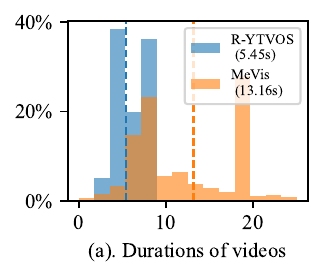}
    \hspace{-3mm}
    \includegraphics[width=0.25\textwidth]{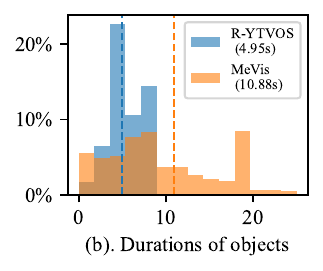}
    \hspace{-4mm}
    \vspace{-3mm}
    \caption{The duration of videos and objects of \ourdataset and Refer-Youtube-VOS~\cite{seo2020urvos}, in seconds. The vertical lines and values in the legends represent the mean duration across the two datasets. The duration of both videos and objects in \ourdataset is significantly longer than Refer-Youtube-VOS.}
    \label{fig:VideoContent2}
    \vspace{-5mm}
\end{figure}

\begin{figure}
    \centering
    \includegraphics[width=0.476\textwidth]{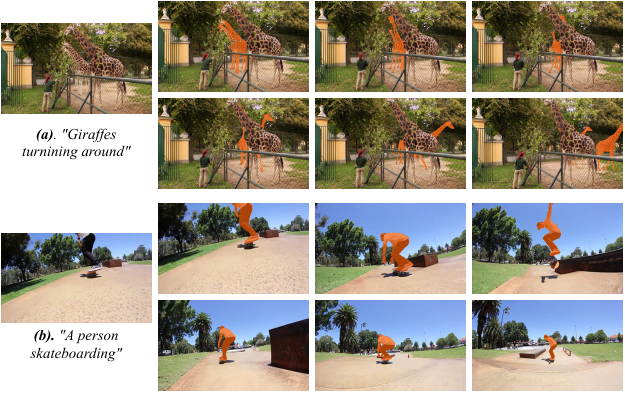}
    \vspace{-7.6mm}
    \caption{(a) Example from \ourdataset. (b) Example from Refer-Youtube-VOS~\cite{seo2020urvos}. Compared to Refer-Youtube-VOS: $\bullet$ Videos in \ourdataset contain \textbf{more objects} in complex environments, making it impossible to identify the target object via saliency or category information alone. $\bullet$ The number of target objects indicated by language expression in \ourdataset is \textbf{arbitrary, from 1 to many}.}
    \label{fig:VideoContent}
    \vspace{-3mm}
\end{figure}

\vspace{0.6mm}
\noindent$\bullet$~\textbf{Video Content.} As shown in \Cref{table:dataset}, \ourdataset has an average of 4.28 objects per video, which is higher than previous datasets. Furthermore, as depicted in \Cref{fig:VideoContent2}, \ourdataset contains longer videos, with an average duration of 13.16 seconds, which is significantly longer than the Refer-Youtube-VOS dataset. These intentional design choices make \ourdataset more complex and challenging for language-guided video segmentation. This is in contrast to existing datasets such as A2D Sentence~\cite{gavrilyuk2018actor} and DAVIS$_{16}$-RVOS~\cite{khoreva2018video}, where only one or two salient objects per category are present, and the model can choose the most prominent object as the target or identify the target object based on the category name. For example, in \Cref{fig:VideoContent}(b), there is only one person in the foreground, and the model can simply identify the target by the term {``\textit{a person}''} while ignoring {``\textit{skateboarding}''}. The proposed \ourdataset dataset addresses this limitation by selecting videos with more objects that have diverse and dynamic motions. Moreover, \ourdataset features many videos with objects of the same category, such as a group of tigers or rabbits. For instance, in \Cref{fig:VideoContent}(a), there are three giraffes with highly similar appearances, and the most salient/foreground one is not the target object in this sample, making it challenging to identify the target object(s) through saliency or category information alone. By including more challenging videos, \ourdataset better simulates real-world scenarios, making it a valuable resource for studying motion expression-guided video understanding in complex environments.

\vspace{1mm}
\noindent\textbf{Target Object(s).} As we have included longer videos in our \ourdataset dataset, we have also observed a significant increase in the duration of target objects, ensuring adequate object motions. As shown in \Cref{fig:VideoContent2}(b), the object durations in our dataset have an average of 10.88 seconds, which is more than two times longer than the average duration of Refer-Youtube-VOS. Compared to previous datasets, such as A2D Sentence~\cite{gavrilyuk2018actor} and J-HMDB Sentence~\cite{gavrilyuk2018actor}, which focus on salient actions of a few categories~\cite{seo2020urvos}, our \ourdataset dataset includes more categories from open-world~\cite{UVO,OVIS,MOSE,TAOVOS}, presenting improved difficulties in the diversity of target objects. Besides, as shown in \Cref{table:dataset}, previous datasets usually have one sentence referring to one single object, which means that finding multiple objects requires multiple expressions, and each object must be searched for individually\footnote{Although some datasets like DAVIS$_{16}$-RVOS~\cite{khoreva2018video} contain samples of multiple instances as targets, the mask annotations provided by DAVIS$_{16}$~\cite{davis2016} do not distinguish instances, and some target masks cover several instances. Thus, each sample in DAVIS$_{16}$-RVOS can be considered as a single-sentence single-object pair.}. In contrast, we add a more natural way of selecting target objects, where one expression may refer to several objects, denoted as ``multi-object expression''. An example of multi-object expression is shown in \Cref{fig:VideoContent}(a), where \textit{``Giraffes that turns around''} refers to two giraffes. As shown in \Cref{table:dataset}, on average, each expression in \ourdataset refers to 1.59 objects, which is larger than existing datasets where the average is only 1 object per expression.

\begin{figure}
    \centering
    \includegraphics[width=0.46\textwidth]{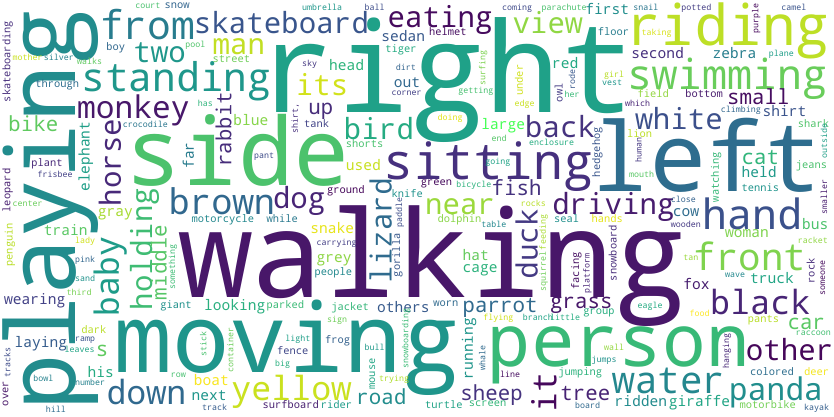}
    \vspace{-2mm}
    \caption{Word cloud of the top 100 words in the \ourdataset dataset. \ourdataset has a large number of words that describe motions, like ``\textit{walking}'',  ``\textit{moving}'', ``\textit{playing}'', and many position words that are related to motions, such as ``\textit{left}'', ``\textit{right}''.}
    \label{fig:wordcloud}
    \vspace{-3.6mm}
\end{figure}

\begin{table*}[htbp]
\centering
\begin{minipage}[t]{1\textwidth}
\scriptsize
 \begin{minipage}[t]{0.56\textwidth}
  \centering
     \small
     \setlength\tabcolsep{3.6pt}
       \caption{Temporal Context ({TC}) shows varying impacts on 3 datasets. Image-based methods, like VLT~\cite{vltpami}, can achieve state-of-the-art performance on DAVIS$_{17}$-RVOS~\cite{khoreva2018video} and Refer-Youtube-VOS (RYV)~\cite{seo2020urvos}, but cannot well handle the harder motion challenges in \ourdataset that require temporal context.}\label{tab:MeViSnecessarity}
  \vspace{-0.05in}
   \setlength{\tabcolsep}{0.96mm}{\begin{tabular}{lccccc}
         \specialrule{.1em}{.05em}{.05em}
         Methods& Type&Temporal& DAVIS$_{17}$-RVOS&RYV & MeViS\\\hline
         VLT~\cite{vltpami}&\textbf{Image}&1 frame&60.4&63.1&27.8\\
         RFormer~\cite{wu2022referformer} &Video&5 rand. frames&60.2&62.8&31.0\\
         VLT+TC&Video&All frames&60.3&62.7&35.5\\
         RFormer+TC&Video&All frames&59.9&63.0&36.3\\
         \specialrule{.1em}{.05em}{.05em}
      \end{tabular}}%
    \vspace{-3mm}
  \end{minipage}
  \hspace{3.6mm}
  \begin{minipage}[t]{0.4\textwidth}
  \small
   \centering
    \caption{Image-video cross-dataset validation. We train the models on referring image segmentation dataset Ref-COCO/+/g, and test their performance on three different video datasets. The models trained on images perform worse on \ourdataset than on the other two datasets.}\label{tab:MeViSnecessarity2}
  \vspace{-0.1in}
  \renewcommand\arraystretch{1.06}
   \setlength{\tabcolsep}{1mm}{\begin{tabular}{lcccc}
         \specialrule{.1em}{.05em}{.05em}
         \rowcolor{gray!10}\multicolumn{5}{c}{Training on Referring Image Segmentation Dataset}\\\hline
         Methods& Type& DAVIS$_{17}$-RVOS&RYV & MeViS\\
         \hline 
         VLT~\cite{vltpami}&Image& 54.2 & 46.1 & 22.5 \\
         RFormer~\cite{wu2022referformer} &Video&55.6&45.2&27.0\\
         \specialrule{.1em}{.05em}{.05em}
      \end{tabular}}%
   \end{minipage}
\end{minipage}
\vspace{-3.6mm}
\end{table*}

\vspace{1mm}
\noindent\textbf{Language Expression.} 
One of the key distinguishing aspects of the \ourdataset dataset is its emphasis on describing object motions in language expressions. The previous largest RVOS dataset Refer-Youtube-VOS~\cite{seo2020urvos} provides two types of language annotations: full-video expression and first-frame expression. The first-frame expression is based solely on static attributes of the first frame image, whereas the full-video expression considers the entire video. However, in many cases, even the full-video expressions contain static attributes that could potentially enable the target object to be identified in a single frame, for example, \textit{``A person on the right dressed in {blue black}...''}. In contrast, to explore the practicality of employing motion expressions for object localization and segmentation in videos, \ourdataset is intentionally designed to include a range of diverse and dynamic object motions, making it more challenging to identify the target object based on static attributes alone. In \ourdataset, there are significantly more motion expressions that explicitly identify the target object based on its distinctive actions or movements. The language expressions in the proposed \ourdataset contain more motion attributes, such as object position moving through the video and actions that span several frames. The word cloud of the newly proposed \ourdataset is visualized in \Cref{fig:wordcloud}. From the word cloud figure, we can observe that \ourdataset dataset has a large number of words that describe motions, like \textit{``walking''}, \textit{``moving''}, \textit{``playing''}, and many relative directions that are related to motions, such as \textit{``left''}, \textit{``right''}, etc.
\section{Experiment}

\myparagraph{Evaluation Metrics.} Similar to previous studies such as~\cite{khoreva2018video,seo2020urvos}, we employ two widely used metrics, \(\mathcal{J}\) and \(\mathcal{F}\), to assess the performance of methods on the newly proposed \ourdataset dataset. The region similarity metric \(\mathcal{J}\) computes the Intersection over Union (IoU) of the predicted and ground-truth masks, which reflects the quality of the segmentation. The F-measure \(\mathcal{F}\) reflects the contour accuracy of the prediction. To provide a comprehensive evaluation of the method's overall effectiveness, we calculate the average of these two metrics, denoted as \(\mathcal{J\&F}\).

\myparagraph{Dataset Setting.} The \ourdataset dataset is a large-scale dataset that consists of a total of \numvideo videos along with \numsentence sentences. 
These videos are split into three subsets, \ie, training set, validation set, and testing set, which contain 1,712 videos, 140 videos, and 154 videos, respectively.

\begin{figure*}
    \centering
    \includegraphics[width=0.999\textwidth]{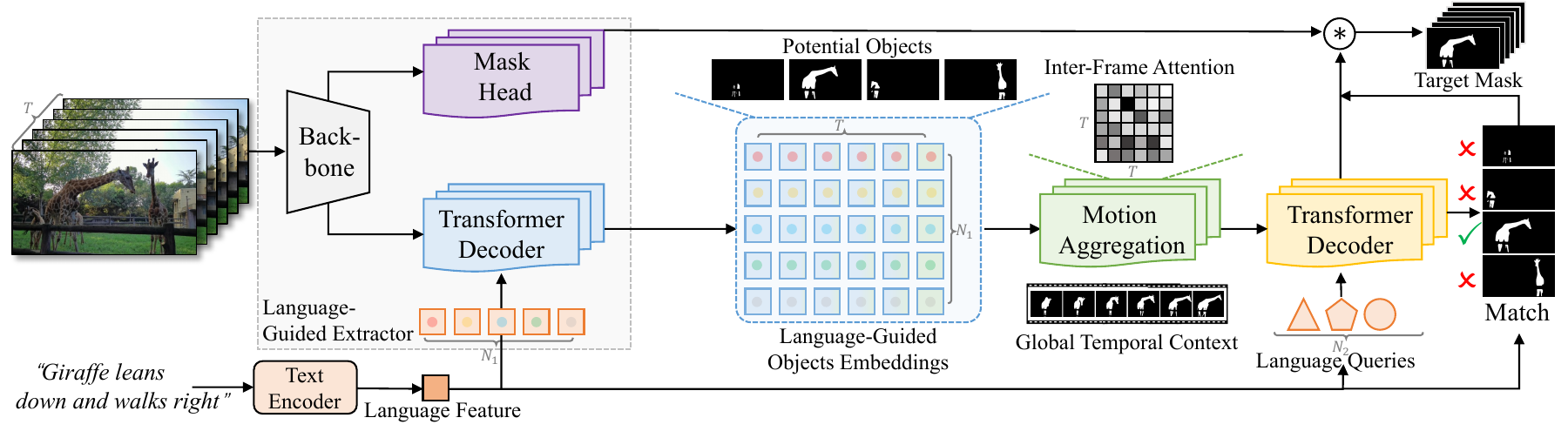}
    \vspace{-5.6mm}
    \caption{The overview architecture of the proposed baseline approach Language-guided Motion Perception and Matching (\textbf{\ourmodel}). We first detect all possible target objects in each frame of the video and use object embeddings to represent them through Language-Guided Extractor. Then, Motion Perception is conducted on all the object embeddings of the video to grasp the global temporal context. By leveraging language queries and object embeddings with motion information, we generate object trajectories through a Transformer Decoder. Finally, we match the language features with the predicted object trajectories to identify the target object(s).}
    \label{fig:L-MAM}
    \vspace{-3mm}
\end{figure*}

\subsection{Dataset Necessity and Challenges}

To show the necessity and validity of \ourdataset in motion expression understanding, we compare the results of state-of-the-art referring image segmentation method VLT~\cite{vltpami} and referring video segmentation method ReferFormer~\cite{wu2022referformer} on DAVIS$_{17}$-RVOS~\cite{khoreva2018video}, Refer-Youtube-VOS~\cite{seo2020urvos}, and \ourdataset, as shown in \Cref{tab:MeViSnecessarity}. When trained on referring video segmentation dataset, such as Refer-Youtube-VOS~\cite{seo2020urvos} and testing on itself, the image-based method VLT~\cite{vltpami} that does not use any temporal design can achieve exceptional results of 60.4\% $\mathcal{J\&F}$ and 63.1\% $\mathcal{J\&F}$ on video datasets DAVIS$_{17}$-RVOS~\cite{khoreva2018video} and Refer-Youtube-VOS~\cite{seo2020urvos}, respectively, which are even better than video method ReferFormer~\cite{wu2022referformer}. The results suggest that for DAVIS$_{17}$-RVOS~\cite{khoreva2018video} and Refer-Youtube-VOS~\cite{seo2020urvos}, the temporal context is not essential, and image-based methods that use static clues can achieve good performance on these two datasets. However, on the proposed \ourdataset, VLT~\cite{vltpami} only achieves a score of 27.8\% $\mathcal{J\&F}$, suggesting that referring image segmentation methods without temporal designs struggle to address the unique challenges presented by videos in our dataset, particularly in handling motion, despite their success on other benchmark datasets. Furthermore, by comparing the results of VLT~\cite{vltpami} with ReferFormer~\cite{wu2022referformer}, which is trained using five randomly selected frames from the video, we find that ReferFormer outperforms VLT by a large margin of {3.2\%} in terms of $\mathcal{J\&F}$. This further highlights the importance of analyzing long-term motions in the \ourdataset dataset. In order to further prove this point, we enhance VLT and ReferFormer by incorporating an attention module at the head to perceive and gather global temporal context (``TC'' in \Cref{tab:MeViSnecessarity}). For module details, please refer to ``Motion Perception'' in \Cref{sec:L-TAM}. Adding temporal context via this module results in both VLT and ReferFormer achieving a performance gain of approximately 5\% $\mathcal{J\&F}$, underscoring the significance of temporal context for \ourdataset. However, it is worth noting that longer temporal information does not necessarily lead to better performance on DAVIS$_{17}$-RVOS and Refer-Youtube-VOS.

We also conduct a cross-dataset experiment by training on referring image segmentation datasets and testing on referring video segmentation datasets. The results in \Cref{tab:MeViSnecessarity2} show that both the image-based method VLT~\cite{vltpami} and video-based method ReferFormer~\cite{wu2022referformer} achieve competitive results on Refer-Youtube-VOS~\cite{seo2020urvos} and DAVIS$_{17}$-RVOS~\cite{khoreva2018video} when trained on image datasets Ref-COCO, Ref-COCO+, and Ref-COCOg. These results suggest that the expressions in Refer-Youtube-VOS~\cite{seo2020urvos} and DAVIS${17}$-RVOS~\cite{khoreva2018video} provide static clues like in the image domain, and many target objects can be identified by examining a single frame solely. In contrast, when trained on referring image segmentation datasets and tested on \ourdataset, both VLT~\cite{vltpami} and ReferFormer~\cite{wu2022referformer} perform worse, indicating that there is a significant expression-gap (\eg, static \vs motion) between \ourdataset and these image domain datasets.

\subsection{\ourmodel: A Simple Baseline Approach}\label{sec:L-TAM}

The \ourdataset dataset introduces unique challenges in detecting and understanding object motions in both video and language contexts. The motions described by language expressions can occur over a random number of frames, making it necessary to capture fleeting actions and movements that occur throughout the entire video. This presents significant challenges for recognizing motions in the video content and the corresponding language expressions. Detecting fleeting actions requires meticulous perceiving of every frame while comprehending complex and extended motion spanning multiple frames requires contextual understanding across the entire duration of the video. Current state-of-the-art methods, such as \cite{wu2022referformer,MTTR,Ding_2022_CVPR}, rely on random sampling of a few frames, which may miss frames containing crucial information described by the given expression. Furthermore, these methods fail to effectively extract temporal contextual information and instead simply use spatial-temporal feature extractors due to the significant burden on computational resources of temporal communication. Additionally, as illustrated in \Cref{sec:MeViS_Dataset}, objects described by language expressions can vary from one to multiple, requiring the output to cover from one to an arbitrary number of objects.

\vspace{-0.5mm}
To address the challenges posed by the \ourdataset dataset, we propose a baseline approach called Language-guided Motion Perception and Matching (\ourmodel), which is depicted in~\Cref{fig:L-MAM}. \ourmodel generates $N_1$ language-based queries to identify potential target objects in the video, across $T$ frames, and produces object embeddings to represent each of them. Using language queries instead of conventional object queries can filter out irrelevant objects and ensure the efficiency and effectiveness of subsequent operations~\cite{ding2021vision,vltpami}. 
Inspired by VITA~\cite{VITA}, we represent objects using object embeddings, which provide instance-specific information, to reduce computational requirements~\cite{li2023transformer, li2023tube}. After obtaining object embeddings from frames in the video, we perform motion perception by inter-frame self-attention on the object embeddings to obtain a global view across T frames. Motion perception enables object embeddings to capture temporal contextual information that spans multiple frames, or even the entire video. Then, we use $N_2$ language queries as the query and the object embeddings after Motion Perception as the key and value for the Transformer decoder. The Transformer decoder decodes language-related information from all object embeddings and aggregates relevant information to predict object trajectories. Finally, we match the language features with the predicted object trajectories to identify the target object(s). Rather than only selecting the best-matched object trajectory, we use a matching threshold $\sigma$ to choose object trajectories only if their similarity with the language features exceeds the threshold $\sigma$. This enables the model to handle not only single-object expressions but also multi-object expressions, which is a unique feature of \ourdataset.

\myparagraph{Implementation Details.} {We set all the hyper-parameters related to the Language-Guided Extractor to the default settings of Mask2Former~\cite{mask2former}, including the backbone, Transformer decoder. We train 150,000 iterations using AdamW optimizer~\cite{loshchilov2017adamw} with a learning rate of 0.00005.
Tiny Swin Transformer~\cite{liu2021swin} is employed as our backbone in all the experiments. The input frames are resized to have a minimum size of 360 pixels on the shorter side and a maximum size of 640 pixels on the longer side, to ensure efficient memory usage on the GPU.
 Motion Perception consists of six layers, and the Transformer decoder employs three layers. For the hyper-parameter settings, we set $\sigma$, $N_1$ and $N_2$ to 0.8, 20, and 10, respectively.
We use RoBERTa~\cite{liu2019roberta} as a text encoder that is consistent with the ReferFormer and is frozen all the time.}

\begin{table}[t]
   \renewcommand\arraystretch{1.06}
   \centering
   \small
   \caption{Ablation study of the baseline approach \ourmodel.}
  \vspace{-0.1in}
   \setlength{\tabcolsep}{1.2mm}{\begin{tabular}{ccccc}
         \specialrule{.1em}{.05em}{.05em}
         ID&Language~Query&Motion Perception & Matching & $\mathcal{J\&F}$ \\
         \hline 
         \romannumeral1&\cmark&\xmarkg&\xmarkg&31.0\\
         \romannumeral2&\cmark&\cmark&\xmarkg&36.3\\
         \romannumeral3&\cmark&\cmark&\cmark&37.2\\
         \specialrule{.1em}{.05em}{.05em}
      \end{tabular}}%
   \label{tab:TAM}%
   \vspace{-5mm}
\end{table}%

\myparagraph{Ablation study of \ourmodel.} 
In \Cref{tab:TAM}, we present an ablation study of the baseline approach \ourmodel. We perform the following three experiments:
(\romannumeral1) First, we use language queries to detect potential target object trajectories and output the best trajectory, similar to ReferFormer~\cite{wu2022referformer}. This variant achieves a $\mathcal{J\&F}$ score of 31.0\%. It relies solely on language information with 5 randomly sampled frames and neglects global temporal context across the video, making it unable to effectively process long-term motions. (\romannumeral2) With the help of Motion Perception, the $\mathcal{J\&F}$ score significantly improves by 5.3\%, as it captures temporal contextual information and a global view of the entire video, which are critical for \ourdataset. (\romannumeral3) Since \ourdataset contains multi-objects expressions, outputting only the object with the highest score is insufficient. We introduce a matching mechanism to identify the target object(s), enabling our method to handle not only single-object expressions but also multi-object expressions. This variant outperforms (\romannumeral2) by 0.9\% in terms of $\mathcal{J\&F}$ score.

\vspace{1mm}
\subsection{\ourdataset Benchmark Results}
\vspace{-2mm}

\myparagraph{Quantitative results.}~We performed a comprehensive evaluation of the \ourdataset dataset to assess the performance of existing methods in the more challenging motion-expression scenarios. We evaluated 1 modified image-based method VLT~\cite{vltpami} and 4 recent state-of-the-art video-based methods, including URVOS~\cite{seo2020urvos}, LBDT~\cite{Ding_2022_CVPR}, MTTR~\cite{MTTR}, and ReferFormer~\cite{wu2022referformer}, on the validation set\footnote{The test set is utilized for evaluation during the competition periods.} of \ourdataset. The evaluation results, presented in \Cref{tab:MeViS}, indicate that the current state-of-the-art methods could only achieve performance ranging from \textbf{27.8\% $\mathcal{J\&F}$} to \textbf{31.0\% $\mathcal{J\&F}$} on the validation set of \ourdataset, while their results on other benchmarks like Refer-Youtube-VOS~\cite{seo2020urvos} and DAVIS$_{17}$-RVOS~\cite{khoreva2018video} are usually above \textbf{60\% $\mathcal{J\&F}$}. Our experiments demonstrate that while notable progress has been made in language-guided video object segmentation on existing benchmarks, the challenges presented by \ourdataset underline the need for further exploration of motion expression-guided video segmentation in complex scenarios. These challenges can arise from various factors, including both linguistic and visual modalities, such as the use of motion expressions and highly dynamic objects or fast-paced motions in videos, which can impact the overall performance of algorithms.

\begin{table}[t]
   \renewcommand\arraystretch{1.0}
   \centering
   \small
   \caption{\ourdataset Benchmark Results.}
  \vspace{-0.12in}
   \setlength{\tabcolsep}{5.2mm}{\begin{tabular}{lccc}
         \specialrule{.1em}{.05em}{.05em}
         Methods&$\mathcal{J\&F}$ & $\mathcal{J}$ & $\mathcal{F}$ \\
         \hline
         URVOS~\cite{seo2020urvos} &27.8&25.7&29.9\\
         LBDT~\cite{Ding_2022_CVPR}&29.3&27.8&30.8\\
         MTTR~\cite{MTTR}&30.0&28.8&31.2\\
         ReferFormer~\cite{wu2022referformer} &31.0&29.8&32.2\\
         VLT+TC~\cite{vltpami}&35.5&33.6&37.3\\
         \hline         
         \textbf{\ourmodel} (ours) & \textbf{37.2} &\textbf{34.2}& \textbf{40.2}\\
         \specialrule{.1em}{.05em}{.05em}
      \end{tabular}}%
   \label{tab:MeViS}%
   \vspace{-0.24in}
\end{table}%

\begin{figure}
    \centering
    \includegraphics[width=\linewidth]{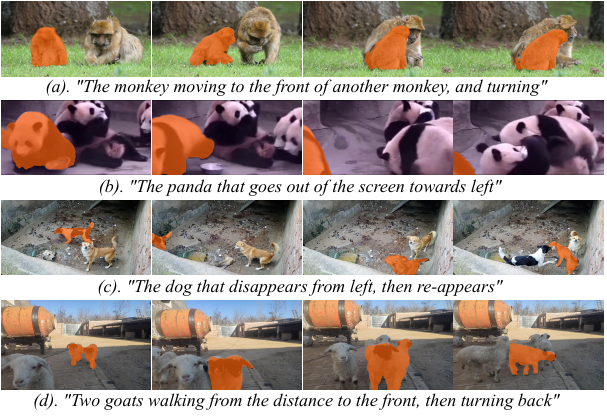}
    \vspace{-6.6mm}
    \caption{Example success and failure cases of \ourmodel.}
    \label{fig:visualizations}
    \vspace{-3mm}
\end{figure}

\myparagraph{Visualizations.} \Cref{fig:visualizations} displays some of the success and failure cases of the baseline approach \ourmodel. Example (a) and (b) depict successful cases where \ourmodel effectively processes expressions of long-term motions such as ``\textit{moving to the front ...}" and ``\textit{goes out of the screen}". In contrast, example (c) and (d) are failure cases. In example (c), the expression involves the target object disappearing and reappearing, which poses a significant challenge for the model's global understanding of the video. In this case, our model becomes disoriented after the target object reappears. Example (d) shows a sentence describing a long-term motion while involving multiple target objects. Our method successfully identifies the multiple targets, \ie, the ``\textit{two goats walking from the distance}", at the beginning of the video. However, one of the targets is lost during the later stages of the video when the motions of objects became complex and tangled. These two failure cases demonstrate the complexity and challenges of \ourdataset, emphasizing the importance of a strong ability to comprehend the global temporal context of the entire video and to understand the motion expression for models working on \ourdataset.

\section{Conclusion and Discussion}

The ability to effectively understand and leverage motion expressions as a primary cue for object segmentation in videos remains an unresolved challenge that requires attention in future research. The proposed large-scale benchmark {\ourdataset} provides a foundation for developing more advanced language-guided video segmentation algorithms.

\textbf{Future Directions.} There are many interesting research directions and remaining challenges to be addressed with the \ourdataset dataset. These include but are not limited to: (\romannumeral1) exploring new techniques for better motion understanding and modeling in both visual and linguistic modalities, (\romannumeral2) designing more elegant and robust models that can effectively handle diverse motion types spanning across a range of frames, including long-term/short-term and complex motions, (\romannumeral3) developing advanced models that can handle complex scenes with various types of objects and expressions, (\romannumeral4) creating more efficient models that can effectively reduce the number of redundant detected objects, (\romannumeral5) designing effective cross-modal fusion methods to better leverage the complementary information between language and visual signals, (\romannumeral6) investigating the potential of transfer learning and domain adaptation in language-guided video segmentation, and (\romannumeral7) developing methods that can better handle the open-world concepts in both the visual and linguistic domain. These challenges require significant research efforts to advance the state-of-the-art in language-guided video segmentation.

{\footnotesize
\bibliographystyle{ieee_fullname}
\bibliography{egbib}

\begin{thebibliography}{10}\itemsep=-1pt

\bibitem{bellver2020refvos}
Miriam Bellver, Carles Ventura, Carina Silberer, Ioannis Kazakos, Jordi Torres,
  and Xavier Giro-i Nieto.
\newblock A closer look at referring expressions for video object segmentation.
\newblock {\em Multimedia Tools and Applications}, 2022.

\bibitem{MTTR}
Adam Botach, Evgenii Zheltonozhskii, and Chaim Baskin.
\newblock End-to-end referring video object segmentation with multimodal
  transformers.
\newblock In {\em Proc. IEEE Conf. Comput. Vis. Pattern Recognit.}, 2022.

\bibitem{chen2022multi}
Weidong Chen, Dexiang Hong, Yuankai Qi, Zhenjun Han, Shuhui Wang, Laiyun Qing,
  Qingming Huang, and Guorong Li.
\newblock Multi-attention network for compressed video referring object
  segmentation.
\newblock In {\em ACM Int. Conf. Multimedia}, 2022.

\bibitem{mask2former}
Bowen Cheng, Ishan Misra, Alexander~G Schwing, Alexander Kirillov, and Rohit
  Girdhar.
\newblock Masked-attention mask transformer for universal image segmentation.
\newblock In {\em Proc. IEEE Conf. Comput. Vis. Pattern Recognit.}, 2022.

\bibitem{ding2020phraseclick}
Henghui Ding, Scott Cohen, Brian Price, and Xudong Jiang.
\newblock {PhraseClick}: toward achieving flexible interactive segmentation by
  phrase and click.
\newblock In {\em Proc. Eur. Conf. Comput. Vis.}, 2020.

\bibitem{BFP}
Henghui Ding, Xudong Jiang, Ai~Qun Liu, Nadia~Magnenat Thalmann, and Gang Wang.
\newblock Boundary-aware feature propagation for scene segmentation.
\newblock In {\em Proc. IEEE Int. Conf. Comput. Vis.}, 2019.

\bibitem{ding2018context}
Henghui Ding, Xudong Jiang, Bing Shuai, Ai~Qun Liu, and Gang Wang.
\newblock Context contrasted feature and gated multi-scale aggregation for
  scene segmentation.
\newblock In {\em Proc. IEEE Conf. Comput. Vis. Pattern Recognit.}, 2018.

\bibitem{MOSE}
Henghui Ding, Chang Liu, Shuting He, Xudong Jiang, Philip~HS Torr, and Song
  Bai.
\newblock {MOSE}: A new dataset for video object segmentation in complex
  scenes.
\newblock In {\em Proc. IEEE Int. Conf. Comput. Vis.}, 2023.

\bibitem{ding2021vision}
Henghui Ding, Chang Liu, Suchen Wang, and Xudong Jiang.
\newblock Vision-language transformer and query generation for referring
  segmentation.
\newblock In {\em Proc. IEEE Int. Conf. Comput. Vis.}, 2021.

\bibitem{vltpami}
Henghui Ding, Chang Liu, Suchen Wang, and Xudong Jiang.
\newblock {VLT}: Vision-language transformer and query generation for referring
  segmentation.
\newblock {\em IEEE Trans. Pattern Anal. Mach. Intell.}, 2023.

\bibitem{Ding_2022_CVPR}
Zihan Ding, Tianrui Hui, Junshi Huang, Xiaoming Wei, Jizhong Han, and Si Liu.
\newblock Language-bridged spatial-temporal interaction for referring video
  object segmentation.
\newblock In {\em Proc. IEEE Conf. Comput. Vis. Pattern Recognit.}, 2022.

\bibitem{feng2021encoder}
Guang Feng, Zhiwei Hu, Lihe Zhang, and Huchuan Lu.
\newblock Encoder fusion network with co-attention embedding for referring
  image segmentation.
\newblock In {\em Proc. IEEE Conf. Comput. Vis. Pattern Recognit.}, 2021.

\bibitem{gavrilyuk2018actor}
Kirill Gavrilyuk, Amir Ghodrati, Zhenyang Li, and Cees~GM Snoek.
\newblock Actor and action video segmentation from a sentence.
\newblock In {\em Proc. IEEE Conf. Comput. Vis. Pattern Recognit.}, 2018.

\bibitem{he2017mask}
Kaiming He, Georgia Gkioxari, Piotr Doll{\'a}r, and Ross Girshick.
\newblock Mask r-cnn.
\newblock In {\em Proc. IEEE Int. Conf. Comput. Vis.}, 2017.

\bibitem{VITA}
Miran Heo, Sukjun Hwang, Seoung~Wug Oh, Joon-Young Lee, and Seon~Joo Kim.
\newblock Vita: Video instance segmentation via object token association.
\newblock In {\em Proc. Adv. Neural Inform. Process. Syst.}, 2022.

\bibitem{hu2016segmentation}
Ronghang Hu, Marcus Rohrbach, and Trevor Darrell.
\newblock Segmentation from natural language expressions.
\newblock In {\em Proc. Eur. Conf. Comput. Vis.}, 2016.

\bibitem{hui2021collaborative}
Tianrui Hui, Shaofei Huang, Si Liu, Zihan Ding, Guanbin Li, Wenguan Wang,
  Jizhong Han, and Fei Wang.
\newblock Collaborative spatial-temporal modeling for language-queried video
  actor segmentation.
\newblock In {\em Proc. IEEE Conf. Comput. Vis. Pattern Recognit.}, 2021.

\bibitem{hui2020linguistic}
Tianrui Hui, Si Liu, Shaofei Huang, Guanbin Li, Sansi Yu, Faxi Zhang, and
  Jizhong Han.
\newblock Linguistic structure guided context modeling for referring image
  segmentation.
\newblock In {\em Proc. Eur. Conf. Comput. Vis.}, 2020.

\bibitem{jing2021locate}
Ya Jing, Tao Kong, Wei Wang, Liang Wang, Lei Li, and Tieniu Tan.
\newblock Locate then segment: A strong pipeline for referring image
  segmentation.
\newblock In {\em Proc. IEEE Conf. Comput. Vis. Pattern Recognit.}, 2021.

\bibitem{kazemzadeh-etal-2014-referitgame}
Sahar Kazemzadeh, Vicente Ordonez, Mark Matten, and Tamara Berg.
\newblock {R}efer{I}t{G}ame: Referring to objects in photographs of natural
  scenes.
\newblock In {\em {Proc. of the Conf. on Empirical Methods in Natural Language
  Process.}}, Doha, Qatar, 2014. Association for Computational Linguistics.

\bibitem{khoreva2018video}
Anna Khoreva, Anna Rohrbach, and Bernt Schiele.
\newblock Video object segmentation with language referring expressions.
\newblock In {\em Proc. Asi. Conf. Comput. Vis.}, 2018.

\bibitem{kim2022restr}
Namyup Kim, Dongwon Kim, Cuiling Lan, Wenjun Zeng, and Suha Kwak.
\newblock Restr: Convolution-free referring image segmentation using
  transformers.
\newblock In {\em Proc. IEEE Conf. Comput. Vis. Pattern Recognit.}, 2022.

\bibitem{li2018referring}
Ruiyu Li, Kaican Li, Yi-Chun Kuo, Michelle Shu, Xiaojuan Qi, Xiaoyong Shen, and
  Jiaya Jia.
\newblock Referring image segmentation via recurrent refinement networks.
\newblock In {\em Proc. IEEE Conf. Comput. Vis. Pattern Recognit.}, 2018.

\bibitem{li2023transformer}
Xiangtai Li, Henghui Ding, Wenwei Zhang, Haobo Yuan, Jiangmiao Pang, Guangliang
  Cheng, Kai Chen, Ziwei Liu, and Chen~Change Loy.
\newblock Transformer-based visual segmentation: A survey.
\newblock {\em arXiv preprint arXiv:2304.09854}, 2023.

\bibitem{li2023tube}
Xiangtai Li, Haobo Yuan, Wenwei Zhang, Guangliang Cheng, Jiangmiao Pang, and
  Chen~Change Loy.
\newblock Tube-link: A flexible cross tube baseline for universal video
  segmentation.
\newblock {\em ICCV}, 2023.

\bibitem{liang2021topdown}
Chen Liang, Yu Wu, Tianfei Zhou, Wenguan Wang, Zongxin Yang, Yunchao Wei, and
  Yi Yang.
\newblock Rethinking cross-modal interaction from a top-down perspective for
  referring video object segmentation.
\newblock {\em arXiv preprint arXiv:2106.01061}, 2021.

\bibitem{GRES}
Chang Liu, Henghui Ding, and Xudong Jiang.
\newblock {GRES}: Generalized referring expression segmentation.
\newblock In {\em Proc. IEEE Conf. Comput. Vis. Pattern Recognit.}, 2023.

\bibitem{M3Att}
Chang Liu, Henghui Ding, Yulun Zhang, and Xudong Jiang.
\newblock Multi-modal mutual attention and iterative interaction for referring
  image segmentation.
\newblock {\em IEEE Trans. Image Processing}, 2023.

\bibitem{ISFP}
Chang Liu, Xudong Jiang, and Henghui Ding.
\newblock Instance-specific feature propagation for referring segmentation.
\newblock {\em IEEE Trans. Multimedia}, 2022.

\bibitem{liu2017recurrent}
Chenxi Liu, Zhe Lin, Xiaohui Shen, Jimei Yang, Xin Lu, and Alan Yuille.
\newblock Recurrent multimodal interaction for referring image segmentation.
\newblock In {\em Proc. IEEE Int. Conf. Comput. Vis.}, 2017.

\bibitem{liu2021cmpc}
Si Liu, Tianrui Hui, Shaofei Huang, Yunchao Wei, Bo Li, and Guanbin Li.
\newblock Cross-modal progressive comprehension for referring segmentation.
\newblock {\em IEEE Trans. Pattern Anal. Mach. Intell.}, 2021.

\bibitem{liu2019roberta}
Yinhan Liu, Myle Ott, Naman Goyal, Jingfei Du, Mandar Joshi, Danqi Chen, Omer
  Levy, Mike Lewis, Luke Zettlemoyer, and Veselin Stoyanov.
\newblock Roberta: A robustly optimized bert pretraining approach.
\newblock {\em arXiv preprint arXiv:1907.11692}, 2019.

\bibitem{liu2021swin}
Ze Liu, Yutong Lin, Yue Cao, Han Hu, Yixuan Wei, Zheng Zhang, Stephen Lin, and
  Baining Guo.
\newblock Swin transformer: Hierarchical vision transformer using shifted
  windows.
\newblock In {\em Proc. IEEE Int. Conf. Comput. Vis.}, 2021.

\bibitem{long2015fully}
Jonathan Long, Evan Shelhamer, and Trevor Darrell.
\newblock Fully convolutional networks for semantic segmentation.
\newblock In {\em Proc. IEEE Conf. Comput. Vis. Pattern Recognit.}, 2015.

\bibitem{loshchilov2017adamw}
Ilya Loshchilov and Frank Hutter.
\newblock Decoupled weight decay regularization.
\newblock In {\em Proc. Int. Conf. Learn. Represent.}, 2019.

\bibitem{mao2016generation}
Junhua Mao, Jonathan Huang, Alexander Toshev, Oana Camburu, Alan~L Yuille, and
  Kevin Murphy.
\newblock Generation and comprehension of unambiguous object descriptions.
\newblock In {\em Proc. IEEE Conf. Comput. Vis. Pattern Recognit.}, 2016.

\bibitem{margffoy2018dynamic}
Edgar Margffoy-Tuay, Juan~C P{\'e}rez, Emilio Botero, and Pablo Arbel{\'a}ez.
\newblock Dynamic multimodal instance segmentation guided by natural language
  queries.
\newblock In {\em Proc. Eur. Conf. Comput. Vis.}, 2018.

\bibitem{mcintosh2020visual}
Bruce McIntosh, Kevin Duarte, Yogesh~S Rawat, and Mubarak Shah.
\newblock Visual-textual capsule routing for text-based video segmentation.
\newblock In {\em Proc. IEEE Conf. Comput. Vis. Pattern Recognit.}, 2020.

\bibitem{ningpolar}
Ke Ning, Lingxi Xie, Fei Wu, and Qi Tian.
\newblock Polar relative positional encoding for video-language segmentation.
\newblock In {\em IJCAI}, 2020.

\bibitem{davis2016}
Federico Perazzi, Jordi Pont-Tuset, Brian McWilliams, Luc Van~Gool, Markus
  Gross, and Alexander Sorkine-Hornung.
\newblock A benchmark dataset and evaluation methodology for video object
  segmentation.
\newblock In {\em Proc. IEEE Conf. Comput. Vis. Pattern Recognit.}, 2016.

\bibitem{davis2017}
Jordi Pont-Tuset, Federico Perazzi, Sergi Caelles, Pablo Arbel{\'a}ez, Alex
  Sorkine-Hornung, and Luc Van~Gool.
\newblock The 2017 davis challenge on video object segmentation.
\newblock {\em arXiv preprint arXiv:1704.00675}, 2017.

\bibitem{OVIS}
Jiyang Qi, Yan Gao, Yao Hu, Xinggang Wang, Xiaoyu Liu, Xiang Bai, Serge
  Belongie, Alan Yuille, Philip~HS Torr, and Song Bai.
\newblock Occluded video instance segmentation: A benchmark.
\newblock {\em Int. J. Comput. Vis.}, 130(8), 2022.

\bibitem{radford2021learning}
Alec Radford, Jong~Wook Kim, Chris Hallacy, Aditya Ramesh, Gabriel Goh,
  Sandhini Agarwal, Girish Sastry, Amanda Askell, Pamela Mishkin, Jack Clark,
  et~al.
\newblock Learning transferable visual models from natural language
  supervision.
\newblock In {\em Proc. Int. Conf. Mach. Learn.}, 2021.

\bibitem{seo2020urvos}
Seonguk Seo, Joon-Young Lee, and Bohyung Han.
\newblock Urvos: Unified referring video object segmentation network with a
  large-scale benchmark.
\newblock In {\em Proc. Eur. Conf. Comput. Vis.}, 2020.

\bibitem{sun2022starting}
Mingjie Sun, Jimin Xiao, Eng~GEE Lim, and Yao Zhao.
\newblock Starting point selection and multiple-standard matching for video
  object segmentation with language annotation.
\newblock {\em IEEE Trans. Multimedia}, 2022.

\bibitem{tang2021human}
Zongheng Tang, Yue Liao, Si Liu, Guanbin Li, Xiaojie Jin, Hongxu Jiang, Qian
  Yu, and Dong Xu.
\newblock Human-centric spatio-temporal video grounding with visual
  transformers.
\newblock {\em IEEE Trans. Circuit Syst. Video Technol.}, 2021.

\bibitem{vaswani2017attention}
Ashish Vaswani, Noam Shazeer, Niki Parmar, Jakob Uszkoreit, Llion Jones,
  Aidan~N Gomez, {\L}ukasz Kaiser, and Illia Polosukhin.
\newblock Attention is all you need.
\newblock In {\em Proc. Adv. Neural Inform. Process. Syst.}, 2017.

\bibitem{TAOVOS}
Paul Voigtlaender, Lishu Luo, Chun Yuan, Yong Jiang, and Bastian Leibe.
\newblock Reducing the annotation effort for video object segmentation
  datasets.
\newblock In {\em Proc. IEEE Winter Conf. Appl. Comput. Vis.}, 2021.

\bibitem{wang2020context}
Hao Wang, Cheng Deng, Fan Ma, and Yi Yang.
\newblock Context modulated dynamic networks for actor and action video
  segmentation with language queries.
\newblock In {\em AAAI}, 2020.

\bibitem{wang2019asymmetric}
Hao Wang, Cheng Deng, Junchi Yan, and Dacheng Tao.
\newblock Asymmetric cross-guided attention network for actor and action video
  segmentation from natural language query.
\newblock In {\em Proc. IEEE Int. Conf. Comput. Vis.}, 2019.

\bibitem{UVO}
Weiyao Wang, Matt Feiszli, Heng Wang, and Du Tran.
\newblock Unidentified video objects: A benchmark for dense, open-world
  segmentation.
\newblock In {\em Proc. IEEE Int. Conf. Comput. Vis.}, 2021.

\bibitem{wang2022cris}
Zhaoqing Wang, Yu Lu, Qiang Li, Xunqiang Tao, Yandong Guo, Mingming Gong, and
  Tongliang Liu.
\newblock Cris: Clip-driven referring image segmentation.
\newblock In {\em Proc. IEEE Conf. Comput. Vis. Pattern Recognit.}, 2022.

\bibitem{wu2020phrasecut}
Chenyun Wu, Zhe Lin, Scott Cohen, Trung Bui, and Subhransu Maji.
\newblock Phrasecut: Language-based image segmentation in the wild.
\newblock In {\em Proc. IEEE Conf. Comput. Vis. Pattern Recognit.}, 2020.

\bibitem{Wu_2022_CVPR}
Dongming Wu, Xingping Dong, Ling Shao, and Jianbing Shen.
\newblock Multi-level representation learning with semantic alignment for
  referring video object segmentation.
\newblock In {\em Proc. IEEE Conf. Comput. Vis. Pattern Recognit.}, 2022.

\bibitem{wu2022referformer}
Jiannan Wu, Yi Jiang, Peize Sun, Zehuan Yuan, and Ping Luo.
\newblock Language as queries for referring video object segmentation.
\newblock In {\em Proc. IEEE Conf. Comput. Vis. Pattern Recognit.}, 2022.

\bibitem{xu2018youtube}
Ning Xu, Linjie Yang, Yuchen Fan, Dingcheng Yue, Yuchen Liang, Jianchao Yang,
  and Thomas Huang.
\newblock Youtube-vos: A large-scale video object segmentation benchmark.
\newblock {\em arXiv preprint arXiv:1809.03327}, 2018.

\bibitem{yang2022tubedetr}
Antoine Yang, Antoine Miech, Josef Sivic, Ivan Laptev, and Cordelia Schmid.
\newblock Tubedetr: Spatio-temporal video grounding with transformers.
\newblock In {\em Proc. IEEE Conf. Comput. Vis. Pattern Recognit.}, 2022.

\bibitem{yang2021bottom}
Sibei Yang, Meng Xia, Guanbin Li, Hong-Yu Zhou, and Yizhou Yu.
\newblock Bottom-up shift and reasoning for referring image segmentation.
\newblock In {\em Proc. IEEE Conf. Comput. Vis. Pattern Recognit.}, 2021.

\bibitem{yang2021lavt}
Zhao Yang, Jiaqi Wang, Yansong Tang, Kai Chen, Hengshuang Zhao, and Philip~HS
  Torr.
\newblock Lavt: Language-aware vision transformer for referring image
  segmentation.
\newblock In {\em Proc. IEEE Conf. Comput. Vis. Pattern Recognit.}, 2022.

\bibitem{ye2019cross}
Linwei Ye, Mrigank Rochan, Zhi Liu, and Yang Wang.
\newblock Cross-modal self-attention network for referring image segmentation.
\newblock In {\em Proc. IEEE Conf. Comput. Vis. Pattern Recognit.}, 2019.

\bibitem{yu2018mattnet}
Licheng Yu, Zhe Lin, Xiaohui Shen, Jimei Yang, Xin Lu, Mohit Bansal, and
  Tamara~L Berg.
\newblock Mattnet: Modular attention network for referring expression
  comprehension.
\newblock In {\em Proc. IEEE Conf. Comput. Vis. Pattern Recognit.}, 2018.

\bibitem{yu2016modeling}
Licheng Yu, Patrick Poirson, Shan Yang, Alexander~C Berg, and Tamara~L Berg.
\newblock Modeling context in referring expressions.
\newblock In {\em Proc. Eur. Conf. Comput. Vis.}, 2016.

\bibitem{zhao2022modeling}
Wangbo Zhao, Kai Wang, Xiangxiang Chu, Fuzhao Xue, Xinchao Wang, and Yang You.
\newblock Modeling motion with multi-modal features for text-based video
  segmentation.
\newblock In {\em Proc. IEEE Conf. Comput. Vis. Pattern Recognit.}, 2022.

\end{thebibliography}
}

\end{document}